%% file: root.tex
\documentclass[conference,letterpaper]{IEEEtran}
\IEEEoverridecommandlockouts

\input{headers}
\IEEEoverridecommandlockouts                              

\usepackage{hyperref}
\hypersetup{
    colorlinks=true,
    citecolor=blue,
    linkcolor=blue,
    filecolor=magenta,      
    urlcolor=gray
    }

    \title{\LARGE \bf
Voxeland: Probabilistic Instance-Aware Semantic Mapping with Evidence-based Uncertainty Quantification*
}

\author{Jose-Luis Matez-Bandera$^{1}$, Pepe Ojeda$^{1}$, Javier Monroy$^{1}$,\\Javier Gonzalez-Jimenez$^{1}$ and Jose-Raul Ruiz-Sarmiento$^{1}$
\thanks{This article was accepted for publication on Robotics and Autonomous Systems. See the published version: \url{https://dx.doi.org/10.1016/j.robot.2026.105668}}
\thanks{$^{1}$The authors are with the Machine Perception and Intelligent Robotics (MAPIR) Group, Malaga Institute for Mechatronics Engineering and Cyber-Physical Systems (IMECH.UMA), University of Malaga, Campus de Teatinos, 29071 Malaga, Spain (e-mail: \href{mailto:josematez@uma.es}{josematez@uma.es};  \href{mailto:ojedamorala@uma.es}{ojedamorala@uma.es}; \href{mailto:jgmonroy@uma.es}{jgmonroy@uma.es};
\href{mailto:javiergonzalez@uma.es}{javiergonzalez@uma.es}; \href{mailto:jotaraul@uma.es}{jotaraul@uma.es})}
}

\begin{document}
\maketitle
\thispagestyle{empty}
\pagestyle{empty}


\begin{abstract}

Robots in human-centered environments require accurate scene understanding to perform high-level tasks effectively. 
This understanding can be achieved through instance-aware semantic mapping, which involves reconstructing elements at the level of individual instances. 
Neural networks, the de facto solution for scene understanding, still face limitations such as overconfident incorrect predictions with out-of-distribution objects or generating inaccurate masks.
Placing excessive reliance on these predictions makes the reconstruction susceptible to errors, reducing the robustness of the resulting maps and hampering robot operation. 
In this work, we propose Voxeland, a probabilistic framework for incrementally building instance-aware semantic maps. Inspired by the Theory of Evidence, Voxeland treats neural network predictions as \textit{subjective opinions} regarding map instances at both geometric and semantic levels. These opinions are aggregated over time to form evidence, and are formalized through a probabilistic model. This enables us to quantify uncertainty in the reconstruction process, facilitating the identification of map areas requiring improvement (e.g. reobservation or reclassification). As a possible strategy to exploit this uncertainty quantification, we incorporate a Large Vision-Language Model (LVLM) to perform semantic level disambiguation for instances with high uncertainty.
Results from the standard benchmarking on the publicly available SceneNN dataset demonstrate that Voxeland outperforms state-of-the-art methods, highlighting the benefits of incorporating and leveraging both instance- and semantic-level uncertainties to enhance reconstruction robustness. This is further validated through qualitative and quantitative experiments conducted on the real-world ScanNet dataset.
\end{abstract}


\section{INTRODUCTION}

For mobile robots to effectively operate in complex, human-centered environments such as factories, healthcare facilities, or homes, achieving accurate scene understanding is a critical prerequisite. For example, stocking specific items on designated shelves in a medical supply room requires the robot to discern between different items, understand the layout of shelves, and perceive spatial relationships within its surroundings. This understanding is codified into a semantic knowledge base linked to the geometric model of the environment (\ie a \textit{semantic map})~\cite{ruiz2017_building}, which enables the robot to reason about the location of items, recognize them, and interact by grasping or relocating as needed, and communicate with humans regarding these items.

Among the different methods for building semantic maps, instance-aware approaches~\cite{grinvald2019_volumetric, wang2019_multi, li2020_incremental, mascaro2022_volumetric} address this problem by annotating elements in a reconstructed 3D scene with semantic information such as object categories, functionalities, or physical characteristics. These elements are mapped at the level of individual instances, permitting the distinction between multiple occurrences of the same object type within a scene. Traditionally, this task involves processing a sequence of RGB-D images with known poses, as outlined in previous works~\cite{grinvald2019_volumetric, mascaro2022_volumetric}.
%
\begin{figure}[t!]
    \centering
    \includegraphics[width=1.0\columnwidth]{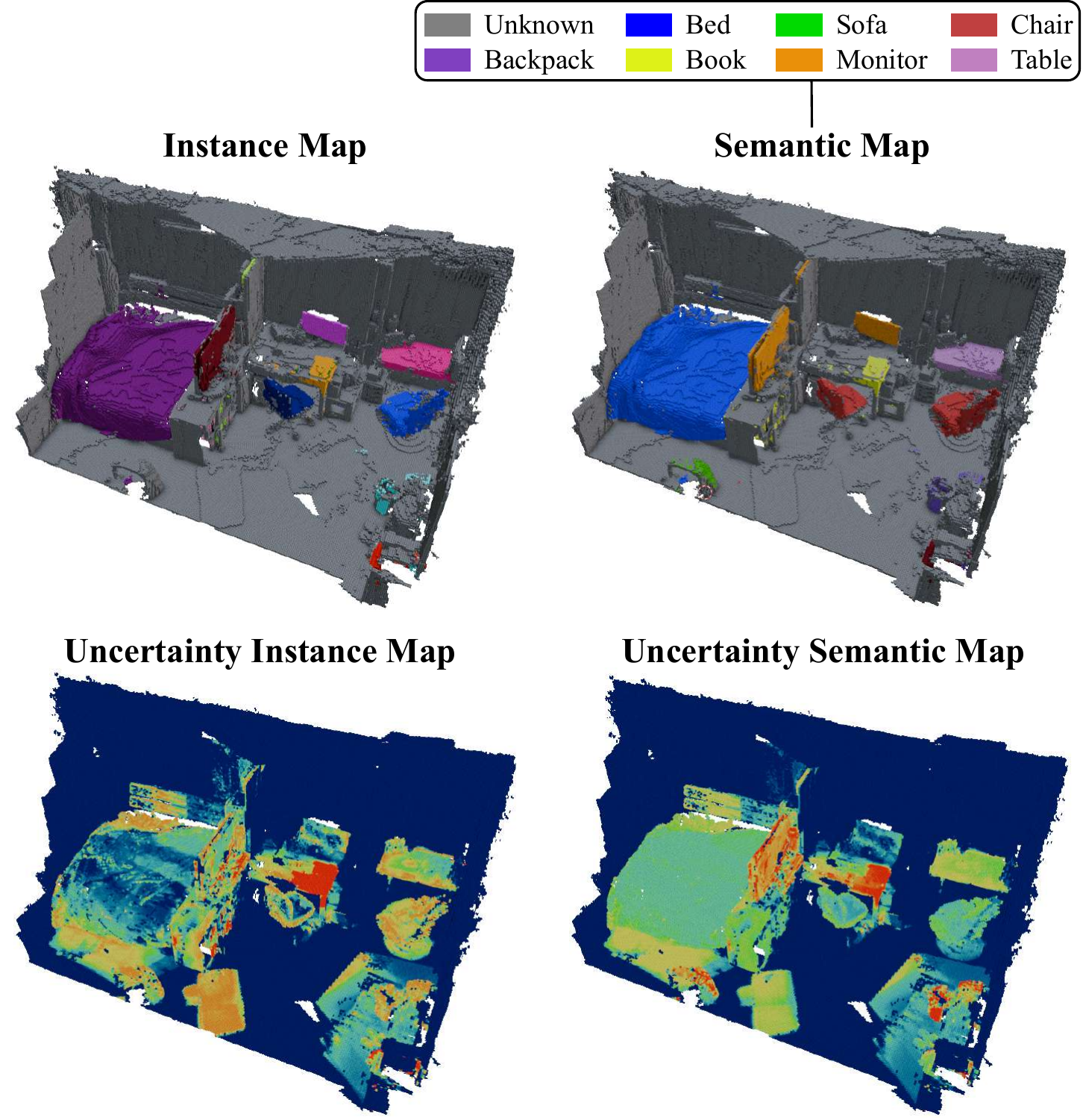}
    \caption{Reconstruction at the instance and semantic levels of the sequence 096 from the SceneNN~\cite{hua2016_scenenn} dataset produced by the proposed method. While each different color in the instance map refers to a different instance, the colors in the semantic map represent the specific category indicated in the top-right legend. Additionally, both generated uncertainty maps are shown, being the blue areas those with minimum uncertainty, while red areas are those with the highest uncertainty. }
    \label{fig:maps}
\end{figure}
Initially, each incoming RGB image is processed using an instance semantic segmentation network that generates object instance predictions, each one labeled with a category and a confidence score. Subsequently, this information is projected into 3D space using the corresponding depth image and is integrated with previous observations through semantic fusion strategies.

Neural networks (NNs) have become the de facto technique for instance semantic segmentation because of their notable perception capabilities. However, when facing out-of-distribution objects (\ie, objects with categories not present in the training data), their performance suffers considerable degradation, leading to erroneous predictions that are often supported by high confidence scores~\cite{morilla2023_robust}. Additionally, the predicted masks frequently exhibit inaccuracies, either including parts of the background or over-segmenting objects~\cite{mascaro2022_volumetric}. Placing excessive trust on the network's predictions makes the semantic map highly prone to these errors, leading to inconsistent representations. Thus, it becomes necessary to incorporate a systematic method for uncertainty quantification into the mapping process.

As is common in many robotic tasks~\cite{matez2022_sigmafp, lee2022_uncertainty}, incorporating uncertainty into the semantic mapping process appears to be a logical and necessary approach.
This uncertainty can be defined across the network's multiple predictions for the same instance at i) the geometric level, to accurately delineate instance boundaries, and ii) at the semantic level, to differentiate between confidently classified objects and those requiring disambiguation.
It must be stressed that state-of-the-art approaches often neglect this uncertainty by adopting simplistic fusion strategies, such as label counting or summing confidence scores~\cite{grinvald2019_volumetric, mascaro2022_volumetric}. These methods typically assign the highest-scored object category to each instance after integrating all the observations, ignoring other possible categories regardless of how close their scores are to the maximum.

In this work, we propose \textit{Voxeland}, a probabilistic framework for the generation of instance-aware semantic maps, coping with uncertainty from the perspective of Evidential Learning~\cite{sensoy2018_evidential} and rooting our formulation in the Dempster–Shafer Theory of Evidence (DST)~\cite{jsang2018_subjective}. Particularly, we interpret each neural network prediction --both the mask and the categorization-- as a subjective opinion, which is accumulated over time to form evidence. 
Our approach handles two types of evidence: i) at the geometric level, where voxels are employed as the representation primitive, and opinions are aggregated to update the belief regarding which object instance a voxel belongs to, and ii) at the semantic level, where each object instance in the map is updated with the incoming opinions about its object category. The integration of these potentially conflicting pieces of evidence into a probabilistic framework is the main contribution of this work, enabling a quantification of both sources of uncertainty, and enhancing the reliability of the resulting semantic maps. 

As illustrated in \FIG{\ref{fig:maps}}, leveraging uncertainty at the geometric level enables us to identify voxels where it is not clear to which instance they belong after multiple conflicting observations. Typically, these conflicts arise from inaccuracies in the 2D prediction masks, or noise in the 3D projection.
Similarly, leveraging uncertainty at the semantic level allows distinguishing which instances may need to be reclassified as being out-of-distribution, or that have received too few observations to be confidently classified. 
As a way to exploit this, instances that require a \textit{disambiguation decision} to discern their final category are evaluated with a Large Vision-Language Model (LVLM), which is provided with multiple observations of the object, the current evidence and its reconstructed geometry.

To implement Voxeland, we build on top of Bonxai~\cite{bonxai2024_faconti}, an open-source library to efficiently store and manipulate volumetric data based on voxel-hashing, which we extend to provide a complete probabilistic framework for instance-aware semantic mapping. Additionally, we rely on probabilistic 3D occupancy mapping with binary Bayes filtering~\cite{thrun2005_probabilistic}, enabling our method to adapt to dynamic scenes. To evaluate our proposal, a set of experiments carried out in multiple sequences from two real-world datasets, SceneNN~\cite{hua2016_scenenn} and ScanNet~\cite{dai2017_scannet}, demonstrates that our approach outperforms state-of-the-art competitors in instance-aware semantic mapping. \FIG{\ref{fig:maps}} illustrates the reconstruction of a scene from SceneNN, as well as the inferred uncertainty maps. 

In summary, our work provides the following contributions:
\begin{itemize}
    \item Voxeland, a complete probabilistic framework for instance-aware semantic mapping rooted in concepts from Theory of Evidence. This allows for the quantification of uncertainty in the predictions integrated into the semantic map.
    \item Generation of uncertainty maps at both geometric and semantic levels, providing valuable insights for identifying areas of the map that require refinement.
    \item An extensive evaluation on the real-world SceneNN dataset, on which main state-of-the-art approaches are evaluated, obtaining an average improvement of $9.8\%$ (Voxeland baseline) and $14\%$ (Voxeland with disambiguation) in instance-level segmentation accuracy. This experimental evaluation is further enhanced with a set of ablation studies. Additionally, the performance of our proposal is further validated with a set of experiments with the ScanNet dataset.
    \item A C++ open-source extension to Bonxai that implements the proposed approach, publicly available as a ROS2 package at: \url{https://github.com/MAPIRlab/Voxeland}.
\end{itemize}

\section{RELATED WORK}

In the context of robotics, semantic mapping refers to the problem of enriching scene geometry maps with high-level information about the properties, functionalities and relations of their constituent elements, such as objects and rooms. This enables the robot to achieve an accurate understanding of the environment, crucial for its interaction with the workspace and its elements. According to how this information is represented in the map, the contributions in this area are divided into two groups: dense approaches and object-oriented approaches.

Dense semantic mapping typically assigns semantic information (e.g., object category) to specific representation models such as points, cells, voxels, or surfels in the map. SemanticFusion~\cite{mccormac2017_semanticfusion} builds a 3D surfel-based map on top of a SLAM system, where per-pixel semantic predictions from a Convolutional Neural Network (CNN) are integrated using a Bayesian filter. Shifting to voxel-based representations, Sun~\etal~\cite{sun2018_recurrent} and Gan~\etal~\cite{gan2020_bayesian} built OctoMap-based models to extend the geometrical information with semantics. Similarly, Diffuser~\cite{mascaro2021_diffuser} propagates and fuses predicted labels into the vertices of a 3D scene mesh by solving a multi-view fusion problem through a graph-based label diffusion approach, exploiting contextual geometry to refine semantics. Morilla-Cabello~\etal~\cite{morilla2023_robust} introduce a robust fusion method that accounts for the network's predictions and their associated uncertainty, enhancing the robustness of the reconstruction. However, a common limitation of all these dense methods is their lack of interpretation of scene objects as individual instances, thus constraining their applicability for robots interacting with their workspace.

In contrast, object-oriented approaches model elements in the environment as individual instances anchored to their corresponding semantics. SLAM++, for instance, proposed mapping scene elements at the instance level, but it operates with a limited number of objects whose geometric models are pre-available in a database. The advent of learning-based models for scene understanding marked a significant breakthrough in this domain, eliminating the need for a pre-existing database of geometric models. Instead, scene objects are directly predicted from RGB images. For example, Sünderhauf~\etal~\cite{sunderhauf2017_meaningful} incrementally build a point cloud of the entire environment, with each point annotated with the instance it belongs to and its object category. Similarly, Voxblox++~\cite{grinvald2019_volumetric} introduces a voxel-based approach for instance-aware semantic mapping that not only detects objects from the set of known categories of the neural networks but also discovers new object-like instances. To refine the geometry of the network's predictions before integrating them into the map, Wang \etal~\cite{wang2019_multi} apply geometric segmentation, while Li~\etal~\cite{li2020_incremental} incorporate a Gaussian Mixture Model (GMM). Alternatively, Mascaro~\etal~\cite{mascaro2022_volumetric} propose a label diffusion scheme that refines the instances geometry directly on the map, removing the need for pre-processing the predictions. Aiming at a more compact representation, Matez-Bandera~\etal~\cite{matez2022_ltc} use 3D bounding boxes as primitives, and introduces the concept of non-detection to explicitly address dynamic changes to improve long-term consistency. Although object-oriented approaches receive more interest because of their direct applicability for high-level robot operation, they often present excessive reliance on neural network's predictions by neglecting their uncertainty, which compromises the resulting map reliability. Our method aligns with these semantic mapping approaches but incorporates a probabilistic framework to enhance the robustness of the resulting maps. 

\section{METHOD}

\begin{figure*}[t!]
    \centering
    \includegraphics[width=1.0\textwidth]{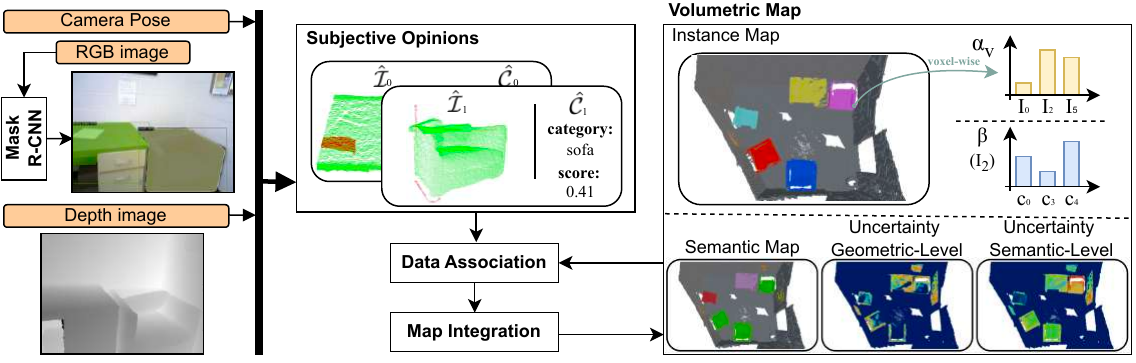}
    \caption{Overview of Voxeland, a probabilistic framework for instance-aware semantic mapping. From a sequence of RGB-D images with known camera poses, our approach processes the RGB images with an instance semantic segmentation neural network and generates per-frame \textit{subjective opinions}. The latter encompasses geometric and semantic levels, which are integrated with the existing map to accumulate evidence. Uncertainty maps at both levels are subsequently derived from their respective accumulated evidence. Please note that, in the uncertainty maps, blue indicates low entropy regions, while red represents high entropy ones.}
    \label{fig:method_overview}
\end{figure*}

An overview of the proposed probabilistic framework, called Voxeland, is illustrated in~\FIG{\ref{fig:method_overview}}, depicting the different phases that compose the pipeline. The framework operates frame-to-frame over an RGB-D sequence to build a volumetric instance-aware semantic map~(\SEC{\ref{subsec:map_representation}}).
Initially, the RGB image is processed by a semantic instance segmentation network, yielding a set of predicted 2D instances within the image. Each instance is annotated with a 2D mask, a category class, and a confidence score. These masks are then projected into 3D space using the depth information, removing points isolated from the rest of the point cloud.
This refined point cloud represents the geometric level \textit{subjective opinion}, while the category class and confidence score constitute the semantic level \textit{subjective opinion}~(\SEC{\ref{subsec:subjective_opinions}}). 
These opinions are subsequently integrated into the semantic mapping framework, which attempts to associate incoming instances with those already existing in the map ~(\SEC{\ref{subsec:data_association}}). If no match is found, a new instance is initiated with the incoming opinion. When an incoming instance matches a map instance, the corresponding voxels are updated with the geometric level opinion, and the map instance is refined by integrating the semantic level opinion~(\SEC{\ref{subsec:integration_and_refinement}}). 
Additionally, Voxeland performs a periodic refinement process to merge map instances that represent the same physical object~(\SEC{\ref{subsec:integration_and_refinement}}). Finally, accumulated opinions at their respective levels form a probabilistic evidence, enabling the quantification and exploitation of uncertainty maps~(\SEC{\ref{subsec:uncertainty_maps}}).

\subsection{Map Representation}
\label{subsec:map_representation}

Our proposal builds on Bonxai~\cite{bonxai2024_faconti}, an efficient library for the sparse representation of volumetric data using voxel-hashing~\cite{niebner2013_realtime}, which extends OpenVDB~\cite{museth2013_vdb}. Voxel-hashing is a widely adopted technique in related works (\eg, Nei{\ss}ner~\etal~\cite{niebner2013_realtime} and Voxblox~\cite{oleynikova2017_voxblox}). The reason for selecting Bonxai over other state-of-the-art voxel-hashing implementations is its explicit probabilistic management of occupancy, which aligns well with our approach.

We start by defining the volumetric map $\mathcal{M}$ as a set of voxels, where each voxel $v$ is annotated with occupancy $\mathcal{O}_v$ and instance $I_v$ probabilities given a set of RGB-D observations $\textbf{z}$.
Concretely, $\text{p}(\mathcal{O}_v | \textbf{z})$ denotes the probability that the voxel is occupied given the observations $\textbf{z}$, computed using a binary Bayesian filter~\cite{thrun2005_probabilistic}, while $\text{p}(I_v | \textbf{z})$ represents the categorical probability distribution over the set of instances in the map $\boldsymbol{\mathcal{I}}$. 
This random variable is defined as a Dirichlet distribution $I_v{\sim}\text{Dir}(\boldsymbol{\alpha}_{v})$, parameterized by a vector $\boldsymbol{\alpha}_{v}=\{\alpha_k \,|\, \mathcal{I}_k \in \boldsymbol{\mathcal{I}} \}$ where $k \in [1,|\boldsymbol{\mathcal{I}}|]$ and $\alpha_k$ represents the number of times a 3D point from the instance $\mathcal{I}_k$ has been registered in that voxel. 
As derived in~\cite{sensoy2018_evidential}, the probability  $\text{p}(I_v = \mathcal{I}_k |\, \textbf{z})$ can be computed as follows:
\begin{equation}
    \text{p}(I_v = \mathcal{I}_k | \,\textbf{z}) = \frac{\alpha_k}{\sum_i\alpha_i}
\end{equation}

It must be noticed that the set of instances in the map $\boldsymbol{\mathcal{I}}$ includes all instances derived from the network’s subjective opinions, plus an additional \textit{unknown} instance. This \textit{unknown} instance accounts for unrecognized scene information, which could be useful for discovering new instances with a different network in the future. 
In turn, each instance $\mathcal{I}_k$ is annotated with p$(C_k | \textbf{z})$ representing the probability of belonging to each possible object category $\mathcal{C}_l \in \boldsymbol{\mathcal{C}}$.
The variable $C_k$ also follows a Dirichlet distribution with concentration parameters $\boldsymbol{\beta}_{k}=\{\beta_l \,|\, \mathcal{C}_l \in \boldsymbol{\mathcal{C}} \}$ where $l \in [1,|\boldsymbol{\mathcal{C}}|]$ and $\beta_l$ represents the accumulated evidence for each category, \ie the sum of confidence scores from various semantic level opinions. 

In summary, each voxel has a probability distribution over the set of instances, and each instance has a probability distribution over the set of object categories.
%

\subsection{Making Subjective Opinions from 2D Semantic Instances}
\label{subsec:subjective_opinions}

For each input RGB-D image, we first apply a semantic instance segmentation network (in our work, Mask R-CNN~\cite{he2017_mask}) to process the RGB image and predict 2D object instances within the image. Each instance prediction comprises a 2D mask, the object category, and a confidence~score.

The predicted masks tend to include pixels that do not belong to the object itself, but rather to adjacent elements or the background. Thus, to refine them, we leverage the corresponding depth information to project the object into 3D space, albeit at a coarse resolution. 
In this coarse 3D space, we assume that all points belonging to the physical object should be connected, \ie, relatively close to each other. We then perform a density-based clustering using DBSCAN~\cite{ester1996_density} to eliminate sparse groups of points that likely do not belong to the object (see subjective opinions in \FIG{\ref{fig:method_overview}}). 

Consequently, each instance within the current image yields a pair of subjective opinions ($\mathcal{\hat{I}}_j$, $\mathcal{\hat{C}}_j$), where $\mathcal{\hat{I}}_j$ are the filtered 3D points depicting the object at the geometric level, and $\mathcal{\hat{C}}_j$ depicts the semantic information, including the category and the confidence score of the prediction. Within our evidential approach, formulating this pair as subjective opinions allows the system to treat the network's predictions as independent evidence rather than absolute probabilistic observations. Specifically for the semantic component ($\mathcal{\hat{C}}_j$), this provides the mechanism for the associated confidence score to continuously update a Dirichlet distribution, explicitly capturing both the expected categorical probability and the underlying epistemic uncertainty of the prediction.


\subsection{Data Association}
\label{subsec:data_association}

To integrate new observations (\textit{subjective opinions}) into the existing map, it is necessary to establish correspondences between the instances within the image and the map instances $\boldsymbol{\mathcal{I}}$. Similar to the proposal in~\cite{mascaro2022_volumetric}, we perform this step at the geometric level in the 3D space, computing the 3D Intersection over Union (IoU) score between each possible pair of a map instance $\mathcal{I}_k$ with an incoming subjective opinion at the geometric level $\mathcal{\hat{I}}_j$:
\begin{equation}
    \text{IoU}(\mathcal{\hat{I}}_j, \mathcal{I}_k) = \frac{\prod(\mathcal{\hat{I}}_j, \mathcal{I}_k)}{|\mathcal{\hat{I}}_j| + |\mathcal{I}_k| - \prod(\mathcal{\hat{I}}_j, \mathcal{I}_k)}, 
\end{equation}
where $\prod(\mathcal{\hat{I}}_j, \mathcal{I}_k)$ refers to the number of 3D points from $\mathcal{\hat{I}}_j$ that lie inside a voxel in which $\mathcal{I}_k$ is mapped, $|\mathcal{\hat{I}}_j|$ represents the number of 3D points in $\mathcal{\hat{I}}_j$, and $|\mathcal{I}_k|$ is the number of voxels in the map instance $\mathcal{I}_k$.

IoU is an effective metric when both sets of data are obtained from similar perspectives, but becomes unreliable when one of the observations is a partial view of the object. 
When comparing an instance that comes from a partial view of a large object to the complete object, the IoU metric yields a low value, incorrectly suggesting that these instances are not from the same object. This issue is relevant to our approach, as partial views of objects are quite common due to the incremental nature of the mapping process. Relying exclusively on IoU to identify instance matches causes frequent false negatives, which leads to over-segmentation. 

To prevent this, we introduce a metric termed \textit{Intersection over Visible} (IoV), which evaluates the spatial overlap relative only to the geometry of the map instance that is currently observable from the camera's viewpoint. The IoV is defined as:
\begin{equation}
    IoV(\mathcal{\hat{I}}_j, \mathcal{I}_k) = \frac{\prod(\mathcal{\hat{I}}_j, \mathcal{I}_k)}{|\mathcal{I}_k^{vis}|}
\end{equation}
where $\mathcal{I}_k^{vis}$ represents the subset of voxels belonging to the map instance $\mathcal{I}_k$ that fall within the current sensor's field of view. By calculating the ratio of the intersecting points to the total number of currently visible points of the map instance, we robustly associate partial views without incorrectly merging distinct, contiguous objects. Thus, $\mathcal{\hat{I}}_j$ is considered to correspond to instance $\mathcal{I}_k$ when $IoU(\mathcal{\hat{I}}_j, \mathcal{I}_k) \ge \tau_{IoU}$ or $IoV(\mathcal{\hat{I}}_j, \mathcal{I}_k) \ge \tau_{IoV}$, where $\tau_{IoU}$ and $\tau_{IoV}$ are empirically set thresholds.

Furthermore, to ensure robustness against uncertainty during this association, we incorporate the \textit{Semantic Similarity} ($\text{S}_{sem}$) between the incoming semantic opinion $\mathcal{\hat{C}}_j$ and the accumulated evidence of the map instance $\mathcal{C}_k$. Given that both are modeled as Dirichlet distributions, we measure their proximity using the Jensen-Shannon Divergence (JSD):
\begin{equation}
    \text{S}_{sem}(\mathcal{\hat{C}}_j, \mathcal{C}_k) = 1 - \text{JSD}(p(\mathcal{\hat{C}}_j|z), p(\mathcal{C}_k|z)).
\end{equation}

This semantic metric allows us to dynamically modulate the geometric integration threshold. If there is high semantic similarity, the IoV threshold is relaxed, allowing the integration of observations with partially noisy geometries but consistent categorizations. Conversely, if semantic uncertainty is high, our proposal requires a more strict geometric match to proceed with the integration.

\subsection{Continuous Map Integration and Refinement}
\label{subsec:integration_and_refinement}

Once the instances from the current RGB-D image are associated with the existing instances in the map, their subjective opinions are integrated to increase the evidence. 

\subsubsection*{\textbf{Geometric Integration}}
At the geometric level, for a pair $(\mathcal{\hat{I}}_j, \mathcal{I}_k)$, we consider each voxel $v$ that contains at least one point from $\mathcal{\hat{I}}_j$. The corresponding concentration parameter $\alpha_k$ for its instance probability variable is updated as follows:
\begin{equation}
    \alpha_{k} \leftarrow \alpha_k + \prod(v, \mathcal{\hat{I}}_j),
\end{equation}
being $\prod(v, \mathcal{\hat{I}}_j)$ the number of 3D points from $\mathcal{\hat{I}}_j$ that lie in voxel $v$. 

It is important to highlight that as the mapping process progresses, additional instances are likely to be detected. In such cases, each new instance expands the set of instances from $\boldsymbol{\mathcal{I}}$ to $\boldsymbol{\mathcal{I}}'$. 
The latter requires re-computing the categorical probability distribution over the set of instances ($\text{p}(I_v | \textbf{z})$) for each voxel in the map, by updating the corresponding concentration parameters $\boldsymbol{\alpha}'_{v}=\{\alpha'_k \,|\, \mathcal{I}_k \in \boldsymbol{\mathcal{I}'} \}$ as:
\[
\alpha'_{k} = 
\begin{cases} 
\alpha_k & \text{if } \mathcal{I}_k \in \boldsymbol{\mathcal{I}} \\
\prod(v, \mathcal{\hat{I}}_j) & \text{otherwise,}
\end{cases}
\]
which in turn results in a new Dirichlet distribution ${I}'_v{\sim}\text{Dir}(\boldsymbol{\alpha}'_{v})$.

\subsubsection*{\textbf{Semantic Integration}}
Concerning the semantic level opinion $\mathcal{\hat{C}}_j$, the integration is performed by updating the concentration parameters $\boldsymbol{\beta}_{k}=\{\beta_l \,|\, \mathcal{C}_l \in \boldsymbol{\mathcal{C}} \}$ of $C_k$ given the previously associated pair $(\mathcal{\hat{I}}_j, \mathcal{I}_k)$. Concretely:
%
%
\begin{equation}
    \beta_l \leftarrow \beta_l + \xi,
\end{equation}
where $\xi$ is the confidence score provided by $\mathcal{\hat{C}}_j$.

Similar to the geometric level, this formulation does not constrain the categories to a closed-set, allowing for extension if new categories are discovered. In this scenario, the addition of new categories results in an extension of the categories set $\boldsymbol{\mathcal{C}}'$, which in turn leads to a new Dirichlet distribution ${C}'_k{\sim}\text{Dir}(\boldsymbol{\beta}'_k)$ whose concentration parameters are set as:
\[
\beta'_{l} = 
\begin{cases} 
\beta_l & \text{if } \mathcal{C}_l \in \boldsymbol{\mathcal{C}} \\
\xi & \text{otherwise,}
\end{cases}
\]
where $\xi$ represents the confidence score of the first prediction of the newly discovered class. 

\subsubsection*{\textbf{Intra-instance Geometric Division}}
While the probabilistic accumulation of evidence naturally diminishes the impact of isolated erroneous predictions, explicit handling of spatial fragmentation is occasionally necessary. This fragmentation typically occurs when an initially under-segmented (overly large) instance is progressively corrected by subsequent, more accurate observations. As new observations correctly claim portions of the geometry for other instances or free space, the original instance is pruned and may become spatially disconnected. To address this, during the periodic refinement process, an intra-instance spatial analysis is performed individually on each map instance. We apply a density-based clustering (DBSCAN) over its belonging voxels. If an instance is found to contain severely disconnected geometries, our method divides the instance into two separate map instances. This ensures that physically separated objects, which were erroneously grouped, are correctly divided into distinct semantic entities.

\subsubsection*{\textbf{Multi-instance Integration}}

Finally, given the nature of incremental instance-aware semantic mapping, non-overlapping parts of the same object (observed from different viewpoints)  might be initially mapped as separate instances. To resolve this over-segmentation, we apply a periodic refinement process (every $N$ frames, empirically set to 30) using a hierarchical decision scheme:
\begin{itemize}
    \item \textbf{Dominant Geometric Match:} If two map instances present an $\text{IoU} > 0.8$, they are directly merged, as they clearly represent the same physical structure.
    \item \textbf{Hybrid Validation:} In cases of moderate overlap, a merge only occurs if both instances simultaneously meet strict \textit{Semantic Similarity} ($\text{S}_{sem}$) and \textit{Intersection over Smaller} (IoS) criteria. Because this comparison is done directly between 3D map entities without a current sensor viewpoint, IoV is not feasible and instead we compute the spatial overlap relative to the smaller instance as:
    \begin{equation}
        \text{IoS}(\mathcal{I}_a, \mathcal{I}_b) = \frac{\prod(\mathcal{I}_a, \mathcal{I}_b)}{\min(|\mathcal{I}_a|, |\mathcal{I}_b|)}
    \end{equation}
    where $\prod(I_a, I_b)$ is the number of intersecting voxels. If either the geometric or semantic check fails, the instances remain distinct.
    \item \textbf{Continuity Verification:} As a final step, a density-based spatial clustering (DBSCAN) is applied over the resulting combined geometry. If the potential merge results in disconnected clusters (e.g., a spatial gap between two objects), the instances are kept as separate entities.
\end{itemize}

Note that, for the purposes of these metrics, the instance geometry includes all the voxels which have at some point been observed as part of the instance, even if there is a different instance which is more likely to contain the voxel given later evidence. The geometric continuity check, on the other hand, considers only the voxels that are currently classified as belonging to the instance (\textit{i.e.} voxels where the instance has received the most votes).

Our current merging strategy fundamentally relies on the existence of at least a moderate geometric intersection between instances. Associating and merging strictly non-overlapping parts of the same physical object, without prior geometric continuity or external reasoning, remains an open challenge and a limitation of this method.

\subsection{Building Uncertainty Maps from Evidence}
\label{subsec:uncertainty_maps}

After incorporating subjective opinions into the map to form evidence, it becomes feasible to quantify the uncertainty linked to the accumulated evidence. This is achieved by computing the expected value of the Shannon entropy for each probability distribution p$(I_v = \mathcal{I}_k| \textbf{z})$ and p$(C_k = \mathcal{C}_l| \textbf{z})$, derived from the corresponding Dirichlet distributions. Specifically, the entropy at the geometric level is calculated voxel-wise by considering the concentration parameters $\boldsymbol{\alpha_{v}}$, as follows:
%
%
\begin{equation}
    H({I}_v) = \psi\left( \sum_{k} \alpha_k \right) - \frac{1}{\sum_{k} \alpha_k} \sum_{k} \alpha_k \psi(\alpha_k),
\end{equation}
where $\psi$(·) denotes the digamma function.
$H({I}_v)$ provides useful insight into the state of the voxel $v$ in the current semantic map, as a high value for this entropy indicates that the current observations are not enough to clearly discern which instance the voxel belongs to, and suggests it might be beneficial to re-observe it.

Similarly, at the semantic level, the entropy is computed by considering the concentration parameters $\boldsymbol{\beta_{k}}$, as:
\begin{equation}
    H({C}_k) = \psi\left( \sum_{l} \beta_l \right) - \frac{1}{\sum_{l} \beta_l} \sum_{l} \beta_l \psi(\beta_l).
\end{equation}

The semantic level uncertainty is useful when declaring a final estimated category for each instance $\mathcal{I}_k$ in the map. That is, while instances with a low entropy can be directly assigned to the class with the maximum probability, those with higher entropy should be further analyzed to discern the correct category.

Finally, a combined metric can be obtained to measure the entropy of the semantic map, \ie the uncertainty about the class of each individual voxel $H(C_v)$. To compute this value, we first apply the Law of Total Probability to obtain the semantic map, \ie the object category distribution per~voxel:
\begin{equation}
   p(C_v| \, \textbf{z}) = \sum_{k} p(C_v |\,\textbf{z}, I_v = \mathcal{I}_{k})\,p(I_v = \mathcal{I}_k |\,\textbf{z}),
\end{equation}
where $p({C}_v |\, \textbf{z}, I_v = \mathcal{I}_k)$ represents the semantic categorization probability of voxel $v$ knowing that it belongs to instance $\mathcal{I}_{k}$. 
Intuitively, this is the same as the probability distribution over the categories for instance $k$, $p(C_k |\,\textbf{z})$. Thus, we can use in this expression the per-instance category probabilities that were previously computed from the Dirichlet distribution.
%

Given that $p(C_v|\,\textbf{z})$ is a discrete distribution (\ie it does not follow a Dirichlet distribution as in the previous cases), the resulting entropy is computed following the particularization of the Shannon equation:
\begin{equation}
    H(C_v) = -\sum_{l} p({C_v = \mathcal{C}_l | \,\textbf{z})}\,\text{log}(p({C_v = \mathcal{C}_l |\,\textbf{z}})).
\end{equation}

\subsection{Uncertainty Exploitation Through Semantic Disambiguation}
\label{subsec:lvlm_exploitation}

{As a strategy to leverage the uncertainty calculated during the mapping process, we implement an offline post-processing semantic disambiguation stage. This process aims to discern the correct category of instances with uncertain classification, which can be identified by the entropy of their probability distribution over the object classes.}


The disambiguation is performed using a Large Vision-Language Model (LVLM), which is fed with comprehensive information about the instance. Specifically, we provide the LVLM with the accumulated evidence p$(C_k = \mathcal{C}_l| \textbf{z})$, the 3D geometry represented through a voxel-based reconstruction, and a collection of images of the object from multiple viewpoints. For the latter, we present $M$ distinct views for each potential category the object might belong to, that is, those with high probability in the evidence (in our case, $M$ is empirically set to 3). For instance, if the object is mainly classified as either a couch or a bed, we provide $M$ views where it was classified as a bed, along with another $M$ views where it was classified as a couch. Next, the LVLM is instructed to produce a \textit{disambiguation decision} based on this information. Besides, the prompt given to the LVLM restricts its response to only potential object categories. In particular, the prompt template is:

\begin{center}
\linespread{0.8}
\footnotesize\texttt{Please, help me to disambiguate the correct category of this object. Here, I provide you with my current evidence (in the form of a probability distribution over the potential categories), its 3D geometry through a voxel-based reconstruction, and a set of views of the object. Given this information, you have to provide an answer in the form of ``The object category is \textbf{{<object\_category>}}'', where only the potential categories provided in the evidence are valid. If none of these categories are correct, reply with the category ``unknown''.}\normalsize
\end{center}


\section{EXPERIMENTS}

\begin{figure*}[t!]
    \centering
    \includegraphics[width=1.0\textwidth]{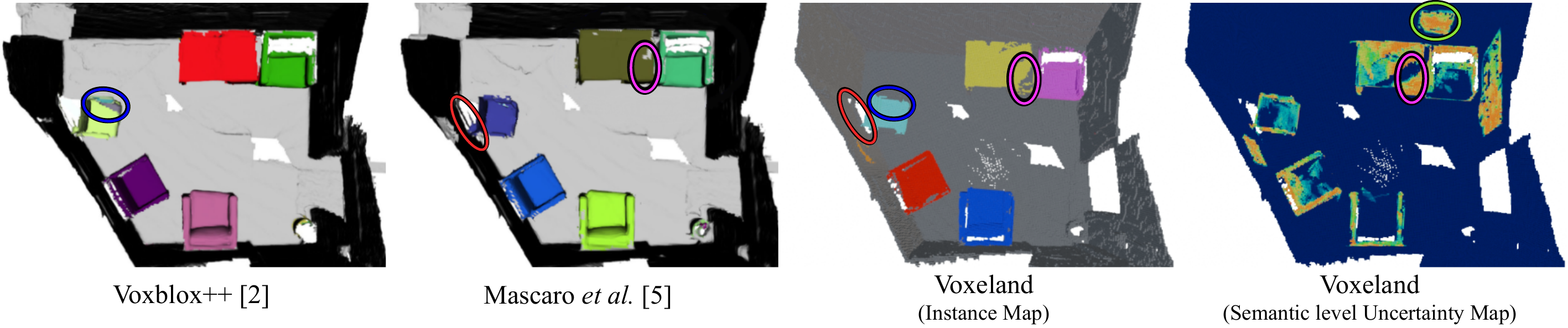}
    \caption{Qualitative results of the instance map for sequence 011 of the SceneNN~\cite{hua2016_scenenn} dataset are shown using two state-of-the-art approaches (Voxblox++~\cite{grinvald2019_volumetric} and Mascaro~\etal~\cite{mascaro2022_volumetric}) and our proposed method. Our method demonstrates a reduction in the over-segmentation problem (highlighted by the blue circle) while improving object delineation without including spurious points (highlighted by the red circle). However, our method exhibits some incomplete areas (highlighted by the magenta circle) due to high uncertainty, specifically because the table was not detected by the neural network in several partial views. Additionally, the green circle in the semantic level uncertainty map indicates an out-of-distribution object (some posters on the wall). For clarification, in the uncertainty map, blue areas represent low entropy while red indicates high entropy regions.}
    \label{fig:qualitative_results}
\end{figure*}

\input{tables/scenenn_eval_rot}

To evaluate the performance of our proposal, we carried out a set of experiments on two real-world datasets, SceneNN~\cite{hua2016_scenenn} and ScanNet~\cite{dai2017_scannet}. On the one hand, the SceneNN dataset is typically employed to benchmark instance-aware semantic mapping methods, thus enabling a comparison with four state-of-the-art approaches~\cite{grinvald2019_volumetric, wang2019_multi, li2020_incremental, mascaro2022_volumetric}. Concretely, the SceneNN dataset provides, for each sequence, RGB-D images of $640\times480\text{ px}$ resolution paired with the camera poses and the ground-truth reconstruction. On the other hand, the ScanNet dataset has been employed to further validate the performance of our method across alternative environments. Similarly to SceneNN, the ScanNet dataset provides posed RGB-D images, as well as a ground-truth mesh reconstruction. Referring to the used object detection technique, for a fair comparison with the competitors, we have resorted to Mask R-CNN~\cite{he2017_mask} with the pre-trained weights for the Microsoft COCO dataset~\cite{lin2014_microsoft} for instance semantic segmentation. All the experiments are conducted on a computer with an Intel i7-11700k CPU and a Nvidia GeForce RTX 3060 Ti GPU with 8 GB of memory. 

For the volumetric map building, we relied on the Bonxai~\cite{bonxai2024_faconti} library, which provides a C++ ROS2 node that already includes probabilistic 3D occupancy mapping~\cite{thrun2005_probabilistic}. We extended Bonxai to implement the proposed probabilistic framework for instance-aware semantic mapping, also providing an extension of the ROS2 node in C++ to handle semantic information. Besides, we developed a separate ROS2 node for the generation of subjective opinions from the RGB-D images, the network's prediction and the camera poses. This node yields a set of subjective opinions at both levels, geometric and semantic, which are the input of the extended Bonxai framework. Finally, to exploit the uncertainty maps, we consider ChatGPT-4o as the LVLM that is used to provide a \textit{disambiguation decision} for the category of instances with high semantic uncertainty. {The entropy threshold to trigger this disambiguation process was experimentally set to $0.8$ nat, based on the observed correlation between the estimation uncertainty and the rate of misclassifications (see Figure~\ref{fig:uncertainty_vs_precision}).} Next, \SEC{\ref{subsec:evaluation_on_SceneNN}} discusses the experimental results, while~\SEC{\ref{subsec:temporal_analysis}} reports the measured execution times for the different phases of the~method. {Section~\ref{sec:ablation} presents a set of ablation studies to analyze the impact of the different aspects of the method. Finally, Section~\ref{sec:scannet} presents a further evaluation on the ScanNet dataset.

\subsection{Evaluation on the SceneNN dataset}
\label{subsec:evaluation_on_SceneNN}

The experiments on the SceneNN dataset~\cite{hua2016_scenenn} are performed in the 10 most commonly used sequences for benchmarking: $011$, $016$, $030$, $061$, $078$, $086$, $096$, $206$, $223$ and $255$. For all the sequences, the voxel size for the reconstruction is set to $5$ cm. The evaluation is performed following the method outlined in~\cite{grinvald2019_volumetric}, considering 9 object categories from the Microsoft COCO dataset~\cite{lin2014_microsoft}: \textit{bed, chair, sofa, table, books, refrigerator, television, toilet and bag}. The metric employed is the per-class Average Precision (AP) score, considering as correct those instances whose 3D representation has an IoU score with the ground-truth greater than or equal to 0.5.

\TABLE{\ref{tab:scenenn_eval}} illustrates the results obtained for the different sequences, showing that our complete proposal outperforms the state-of-the-art, obtaining a 80.8\% mAP on average. In particular, our proposal achieves the best performance in 7 out of 10 sequences. The table also shows the results obtained by Voxeland before exploiting semantic uncertainty (namely, Voxeland w/o Disambiguation). It is noteworthy that even when relying solely on the top-1 classification to assign the final category, our proposal achieves, on average, a 76.6\% mAP, independently outperforming the previous state-of-the-art (Voxblox++~\cite{grinvald2019_volumetric}) by 9.8\%. The results obtained by the full method, however, indicate that leveraging semantic disambiguation provides a further refinement, making the resulting maps more reliable.

It should also be noted that the disambiguation decision technique employed here (asking a separate model for a reclassification of instances with high category uncertainty) is not the full extent of the applicability of the measure of uncertainty. Further research into the topic of how to best exploit this information is necessary, including how to include geometric level uncertainty in the refinement process.

\begin{figure}[t!]
    \centering
    \includegraphics[width=1.0\columnwidth]{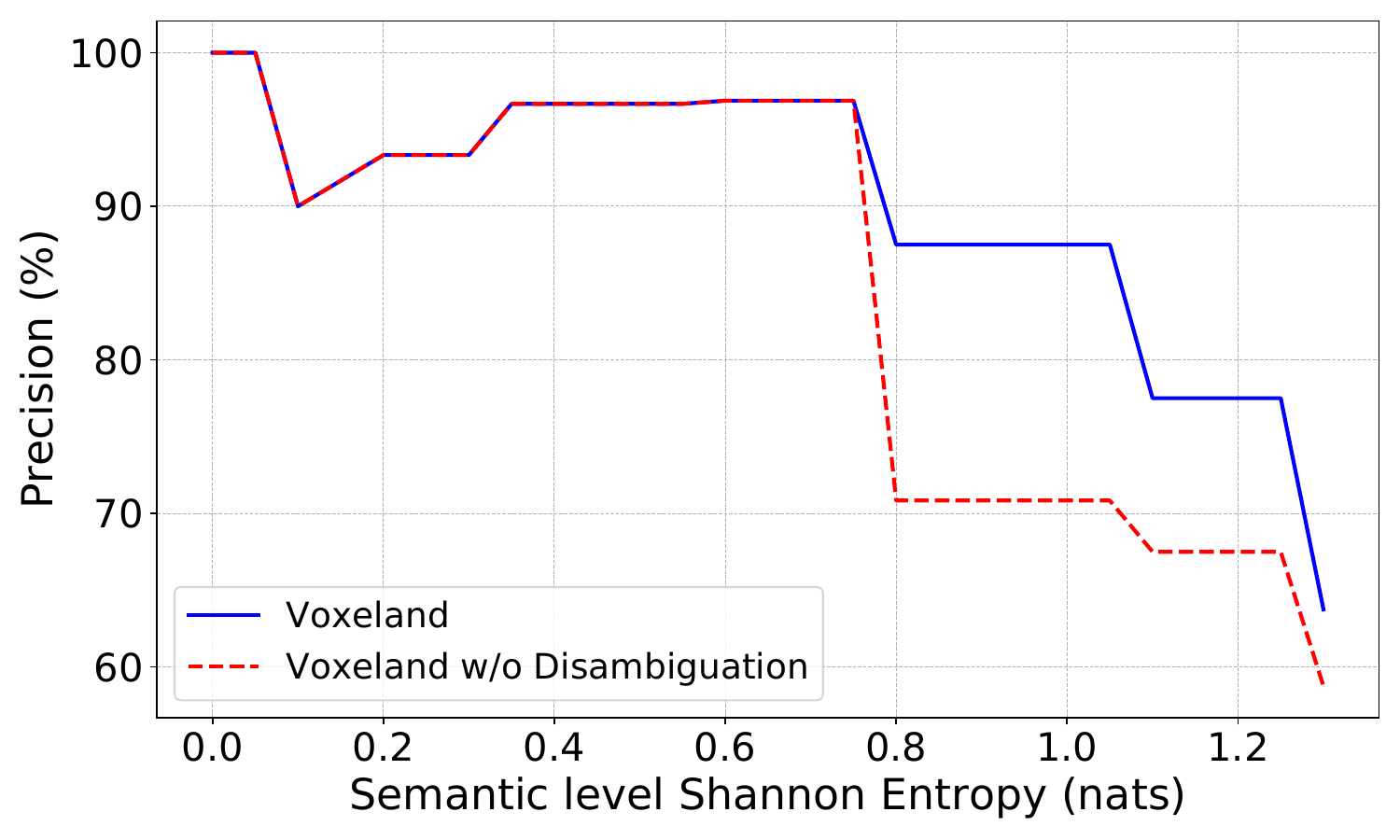}
    \caption{Precision versus semantic level Shannon entropy curves for our proposed method with (blue solid line) and without (red dashed line) uncertainty map exploitation. Precision is calculated for each Shannon entropy value, considering only objects with an associated entropy lower than that threshold. The results indicate that instances with higher semantic level uncertainty often correspond to incorrect categorizations, suggesting that directly assigning the top-1 classification reduces overall performance.}
    \label{fig:uncertainty_vs_precision}
\end{figure}

Additionally, for a better understanding of the importance of considering uncertainty, \FIG{\ref{fig:uncertainty_vs_precision}} shows the precision of our approach with and without semantic disambiguation. This precision is computed for different thresholds of maximum Shannon entropy, \ie only instances with an entropy lower than a certain threshold are considered for the precision. These results show that, in general, those map instances with greater entropy at the semantic level are more likely to be incorrect, in some cases due to being out-of-distribution objects. The effect of considering the uncertainty in those cases is notable, showing that instead of assigning the top-1 category, looking for a more informed second opinion to disambiguate results in a greater robustness in the reconstruction. An example of the latter is shown in~\FIG{\ref{fig:lvlm_disambiguation}}, in which the top-1 classification of a physical couch is \textit{bed} with a 48\%, and the second-best class is \textit{couch} with 46\%. This shows an object with a high semantic entropy associated, in which selecting the top-1 classification would result in an incorrect classification. In contrast, asking the LVLM for a more informed opinion allows for disambiguation, assigning the correct class. Finally, back to~\FIG{\ref{fig:uncertainty_vs_precision}}, it should be noted that still, for both cases, it is appreciable that in the range of low entropy levels ($<$0.4 nats), there is a degradation of the precision, which is the result of some instances which have low entropy at the semantic level, but high at the geometric level, not surpassing the 0.5 threshold in IoU to be considered as correct.

\begin{figure}[t!]
    \centering
    \includegraphics[width=1.0\columnwidth]{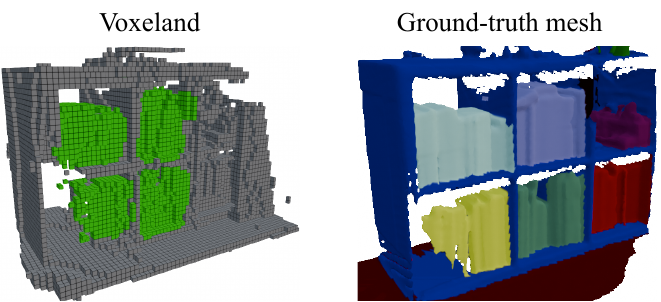}
    \caption{Illustration of the under-segmentation problem in our proposed method. Our approach represents the entire set of books as a single instance (in green), whereas the ground-truth data identifies each group of books per shelf as separate instances.}
    \label{fig:books_problem}
\end{figure}

Finally, observing the results obtained in ~\TABLE{\ref{tab:scenenn_eval}}, it is notable that the smaller objects (\eg books) are those in which the performance of our proposal is reduced. Analyzing the observations in these sequences, we suffer from the same problem as the one mentioned in~\cite{mascaro2022_volumetric}, \ie the masks predicted by the Mask R-CNN are consistently inaccurate, and thus the propagation of the errors to the map is unavoidable. An example can be seen in~\FIG{\ref{fig:books_problem}}, which shows a bookcase with books, where our method under-segments the books, representing all books as a single instance, while the ground-truth separates them by groups of books on each rack. We expect that this problem could be tackled by leveraging the uncertainty at the geometric level.

\begin{figure}[t!]
    \centering
    \includegraphics[width=1.0\columnwidth]{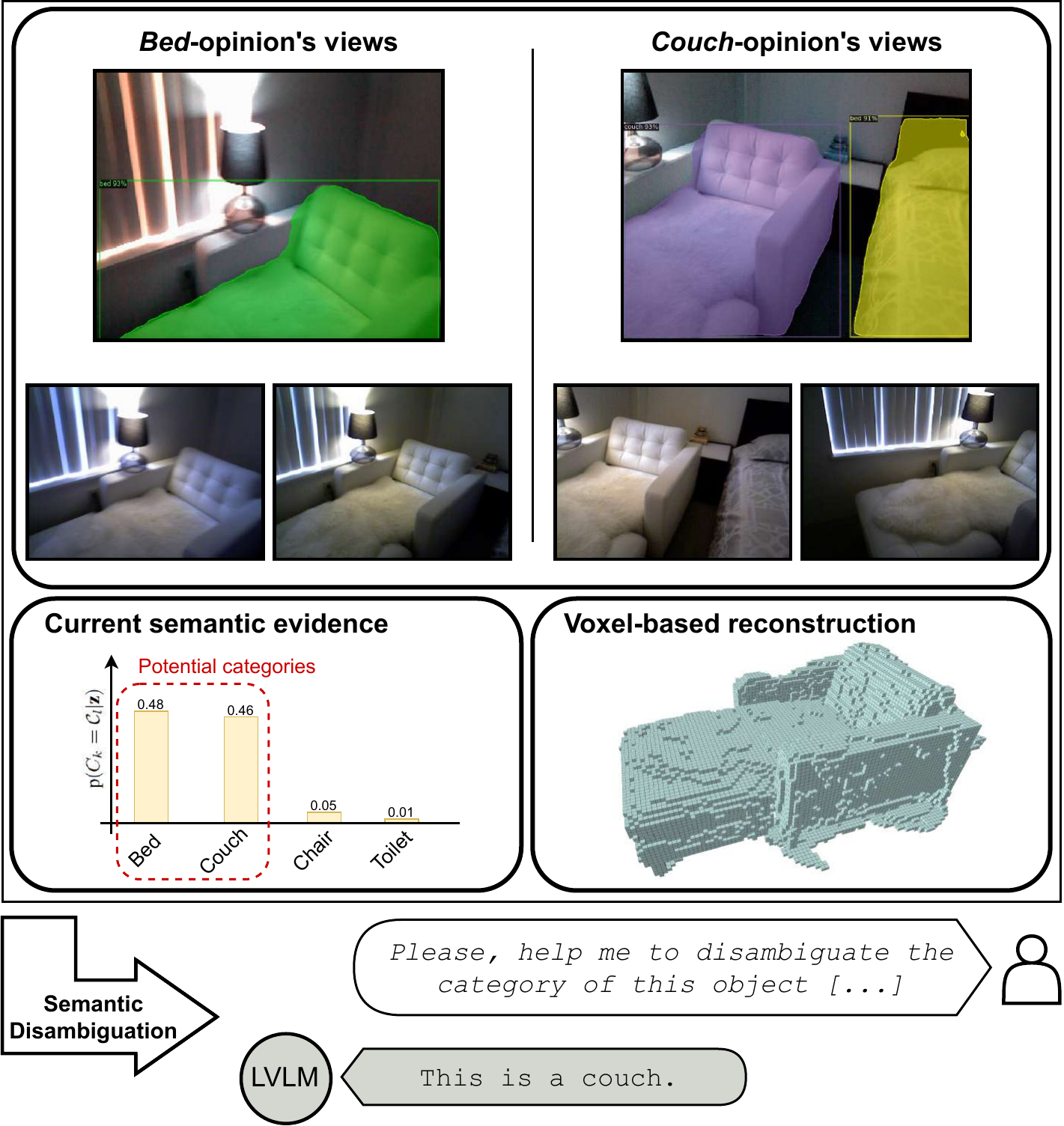}
    \caption{Illustrative example of semantic disambiguation using an LVLM. A certain physical object (\ie a couch) exhibits high semantic entropy due to conflicting category predictions in the accumulated evidence: \textit{bed} and \textit{couch}. While selecting the top-1 prediction (\ie \textit{bed}) would result in misclassification, asking the LVLM for a more informed second opinion enables successful disambiguation, correctly identifying the object as a \textit{couch} by selecting the second-best class.}
    \label{fig:lvlm_disambiguation}
\end{figure}

\subsection{Temporal Analysis of Instance-Aware Semantic Mapping}
\label{subsec:temporal_analysis}

\input{tables/times}
\TABLE{\ref{tab:times}} shows the average execution times for the different stages of Voxeland in the evaluated sequences of the SceneNN~\cite{hua2016_scenenn} dataset. Clearly, the bottleneck comes from the opinions generation, as the 3D point cloud filtering is computationally expensive when dealing with large objects. The latter could be improved in the future by applying parallelization techniques or analyzing the continuity with more efficient volume representations than a regular grid. With the current implementation, our proposal runs at ${\sim}6.24$~Hz, which is enough for online operation of a robot inspecting an environment.

\subsection{Ablation Studies}
\label{sec:ablation}
In order to evaluate the impact of the different aspects of our method, and how relevant their contribution is to the quality of the resulting semantic maps, we have performed a series of ablation studies on the previously discussed sequences of the SceneNN dataset.

Specifically, we have analyzed the performance of our method when removing the following aspects of the proposed method:

\begin{itemize}
    \item 
    \textbf{Point cloud preprocessing:} As mentioned in Section~\ref{subsec:subjective_opinions}, the predictions produced by the instance segmentation network are combined with the depth image to generate a semantically annotated 3D point cloud. This point cloud is then processed to remove outliers and make sure the information that is fed into the map is geometrically coherent. In this study, this preprocessing step is removed.

    \item 
    \textbf{Proposed geometric overlap metrics (IoV and IoS)}: In this study, both the integration of new observations and the refinement of the global map through instance fusion is performed utilizing the IoU as the only geometric similarity metric. As discussed in Section~\ref{subsec:data_association}, the use of IoU can often lead to false negatives when dealing with partial views of large objects, as the intersection is strictly bounded by the portion of the object which is in view. A similar problem occurs when fusing instances in the map if the object was partially observed from different points of view without sufficient overlap.

    \item 
    \textbf{Semantic Similarity metric:} As discussed in Section~\ref{subsec:data_association}, the similarity between the probability distributions over the object categories (measured through the Jensen-Shannon Divergence) is used to relax the geometric overlap requirements when integrating new observations. This same metric is also used as a necessary condition for the fusion of instances in the map during refinement. Removing this metric leads to reliance on fixed thresholds for geometric overlap on both the integration and the refinement phase, which can cause instances to be incorrectly merged in some cases.

    \item 
    \textbf{Geometric clustering refinement:} In the process of map refinement (section~\ref{subsec:integration_and_refinement}), geometric clustering is used for two purposes: to split instances which have become collections of loose pieces, and to ensure the geometric continuity of any pair of instances which meets the rest of the requirements for fusion.

\end{itemize}

\input{tables/Ablation}

Table~\ref{tab:ablation} shows the results for these ablation studies. It can be observed that the two modifications with the greatest impact are the removal of the preprocessing step and removal of the $IoV$ and $IoS$ metrics. In both cases, the degradation of the geometric association between new observations and pre-existing instances leads to an accumulation of errors which cannot be completely corrected by subsequent observations. Removing the semantic similarity metric also leads to a significant reduction in the quality of the resulting maps, as the reliance on a fixed threshold of geometric overlap can cause both under-segmentation and over-segmentation, especially when the masks produced by the instance segmentation network are unreliable. The clustering refinement step has the smallest impact, since employing the rest of the metrics when performing the integration of new observations tends to prevent the inclusion of incorrect instances in the map, which reduces the importance of the refinement step.

\subsection{Evaluation on ScanNet Dataset}
\label{sec:scannet}

\begin{figure*}[t!]
    \centering
    \includegraphics[width=0.9\textwidth]{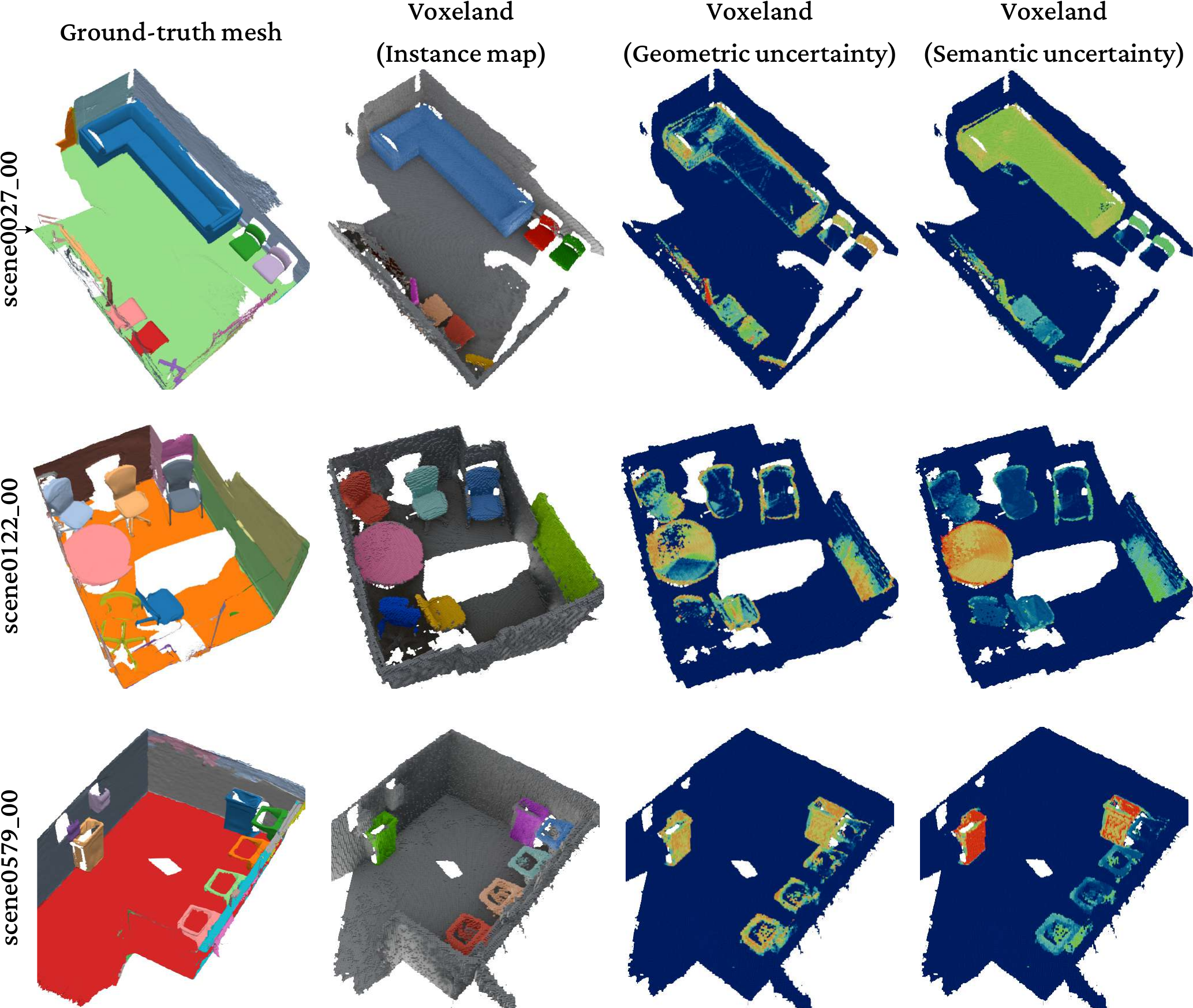}
    \caption{Qualitative results for the ScanNet~\cite{dai2017_scannet} dataset. The instance maps, along with the reconstructed geometric and semantic uncertainty maps using Voxeland are shown. In the uncertainty maps, blue indicates low entropy regions, while red indicates high entropy ones.}
    \label{fig:scannet}
\end{figure*}

In addition to the standard benchmark evaluation discussed in previous sections, for a more comprehensive evaluation in alternative scenarios we also provide quantitative and qualitative results for the ScanNet~\cite{dai2017_scannet} dataset. It is important to note, however, that since there are no results from other state-of-the-art methods for this dataset, we compare just against the ground-truth to demonstrate Voxeland generalization.

\input{tables/scannet_quantitative}

Applying the same evaluation setup used for SceneNN, \TABLE{\ref{tab:scannet_eval}} details the quantitative performance of Voxeland across three different ScanNet sequences. Our framework maintains strong performance, achieving an overall average mAP of 79.2\% without disambiguation, and 95.8\% with disambiguation, across these environments. \FIG{\ref{fig:scannet}} illustrates the resulting reconstructions for these same three scenes. In general, our method effectively captures the instances present in the scene. However, certain elements, such as the two \textit{paper dispensers} located above the orange trash in the ground-truth of \texttt{scene0579\_00}, are not mapped. The latter is due to the fact that these objects are not included in the COCO dataset, and thus the neural network employed for object detection is unable to detect them.

Additionally, it is noticeable that the geometric uncertainty tends to concentrate around the boundaries of the objects and within objects which are challenging for the network to detect. Some of these objects may be properly detected in certain frames but missed in others, often due to challenging viewpoints, as is the case of the table in \texttt{scene0122\_00}. Furthermore, looking at the semantic uncertainty maps, it can be seen that certain objects have been classified as different object classes across various frames, leading to high entropy in their classification. Such is the case, for example, with the couch in \texttt{scene0027\_00}, which is a known case of this network tending to confuse a couch with a bed for some viewpoints, as was also previously illustrated in~\FIG{\ref{fig:lvlm_disambiguation}}. Again, these cases demonstrate the relevance of quantifying the uncertainty, allowing to discern which scene elements may require re-observation (\ie, high geometric uncertainty) or re-classification (\ie, high semantic uncertainty).

\section{CONCLUSIONS}

In this work, we proposed Voxeland, a probabilistic framework that builds instance-aware semantic voxel maps from RGB-D sequences.
Our proposal, rooted in concepts from the Theory of Evidence, interprets neural network predictions as \textit{subjective opinions} at both geometric and semantic levels. The aggregation of these opinions into evidence is formalized through a probabilistic model, setting Voxeland apart from state-of-the-art approaches by allowing us to explicitly quantify and leverage uncertainty. This probabilistic approach provides valuable insights for guiding actions such as reclassification or re-observation, thus enhancing the consistency of the reconstruction.
Results obtained on SceneNN, a publicly available benchmark including real-world sequences, show that our method outperforms state-of-the-art approaches, demonstrating the effectiveness of considering the uncertainty to improve the robustness of the reconstruction. The latter is further validated through a series of qualitative experiments carried out on the ScanNet dataset.

Future work involves the extension of our proposal to exploit the geometric level uncertainty to, for example, improve the reconstruction, reducing the under-segmentation problem. Besides, we plan to incorporate the camera localization uncertainty into the formulation, which is relevant for the deployment of robots in real environments.




\section*{Acknowledgements}
This work has been supported by the grant program FPU19/00704, the research projects MINDMAPS (PID2023-148191NB-I00) and ARPEGGIO (PID2020-117057), both funded by the Spanish Government, and VOXELAND (PPRO-B1-2023-017, JA.B1-09), financed by the University of Malaga, as well as by the Andalusian Plan for Research, Development, and Innovation (PPRO-TEP960-G-2023). Funding for open access charge: Universidad de Málaga / CBUA.

\bibliographystyle{elsarticle-num}
\bibliography{biblio.bib}

\end{document}

%% file: headers.tex
\usepackage{tikz} 
\usepackage[utf8]{inputenc}
\usepackage{mathtools}          
\usepackage{acro}               
\usepackage{comment}
\usepackage{amsmath}
\usepackage{amssymb}
\usepackage[dvipsnames]{xcolor}
\usepackage{bookmark}
\usepackage{tabularx,booktabs}
\usepackage{multirow}
\usepackage[ruled,vlined]{algorithm2e}
\usepackage{doi}
\usepackage{xurl}
\usepackage{lipsum}
\usepackage{cuted} 
\usepackage{xfrac}
\usepackage{resizegather}
\usepackage{rotating} 
\usepackage{ragged2e} 

\providecommand{\etal}{\emph{et al.\,\xspace}}
\providecommand{\ie}{\emph{i.e.\,\xspace}}
\providecommand{\eg}{\emph{e.g.\,\xspace}}

\providecommand{\FIG}[1]{Figure~{#1}}
\providecommand{\FIG}[2]{Figures~{#1} and~{#2}}
\providecommand{\SEC}[1]{Section~{#1}}
\providecommand{\TABLE}[1]{Table~{#1}}

\newcommand\new[1]{\textcolor{black}{#1}}

\def\rot{\rotatebox}

%% file: tables/scenenn_eval_rot.tex
\begin{sidewaystable*}[htp]
\scriptsize
\centering
\centering
\caption{{Comparison of 3D semantic instance-level segmentation performance among four state-of-the-art methods and Voxeland on the SceneNN dataset~\cite{hua2016_scenenn}. On the left, the Average Precision (AP) is shown for each class (calculated with an Intersection over Union (IoU) threshold over 0.5). A dash indicates object category not present in the sequence. The mean Average Precision (mAP) is derived by averaging the per-class AP values. In bold, best results while second-best are underlined.}}
\label{tab:scenenn_eval}
\begin{tabular}{c|ccccccccc||cccccc}
\toprule
\rot{90}{Sequence ID} & \rot{90}{Bed} & \rot{90}{Chair} & \rot{90}{Sofa} & \rot{90}{Table} & \rot{90}{Books} & \rot{90}{Refrigerator} & \rot{90}{Television} & \rot{90}{Toilet} & \rot{90}{Bag} & \rot{90}{Voxeland (Ours)} &  \multicolumn{1}{c}{\rot{90}{\begin{tabular}[c]{@{}l@{}}Voxeland w/o\\Disambiguation\end{tabular}}} & \rot{90}{Voxblox++~\cite{grinvald2019_volumetric}} & \rot{90}{Wang~\etal~\cite{wang2019_multi}} & \rot{90}{Li~\etal~\cite{li2020_incremental}} & \rot{90}{Mascaro~\etal~\cite{mascaro2022_volumetric}} \\ \midrule
011         &  --   &   {100}    &  {100}    &   {100}    &  --     &   --  &   --      &      --      &    --    &  {\textbf{100}}   &   {\textbf{100}} &   75.0       &                   62.2                      &                      \underline{78.6}                                    &                  \textbf{100}                       \\
016         &  {100}   &    {0.0}   &   {100}   &   --    &   --    &      --        &      --      &   --     &  --   &      {\underline{66.7}}    & {33.3}   &                       33.3                  &                     43.0                    &                       25.0                  &               \textbf{75.0}                          \\
030         &  --   &    {100}   &  {100}    &   {66.7}    &   {100}    &      --        &      --      &    --    &  --   &     {\textbf{91.7}}    & {\textbf{91.7}}    &                     56.1                    &                      60.7                   &                   58.6                      &                       \underline{72.5}                  \\ 
061         &  --   &   --    &   {100}   &    {33.3}   &   --    &      --        &     --       &   --     &  --   &      {\underline{66.5}}  & {\underline{66.5}}       &                        \textbf{66.7}                 &                      36.3                   &                     46.6                    &                    50.0                     \\
078      & --  &  {100}   &  --     &   {0.0}   &   {82.0}    &  {50.0}     &       {100}       &    --        &   --     &         {\underline{66.4}}   &            {59.6} &             45.2                &                        49.3                 &                         \textbf{69.8}                &                   50.0                      \\
086         &   -- &    {75.0}  &    --   &   --   &    {50.0}   &   --    &     {100}         &      --      &  {100}      &          {\textbf{81.3}}      &        {\textbf{81.3}} &                20.0                 &                         45.8                &                       47.2                  &               \underline{50.0}                          \\
096         &   {100}  &   {75.0}    &   --   &    {50.0}   &    {0.0}   &    --          &     {100}       &    --    &  {50.0}   &      {\textbf{61.0}}       &   {\underline{58.7}} &               29.2                       &                          32.7               &                    26.7                     &                  51.3                      \\
206         &   --  &   {100}    &  {100}    &   {72.0}    &  --     &     --         &     --       &   --     &  {100}   &      {\textbf{93.0}}       &     {\textbf{93.0}} &               \underline{79.6}                     &                                  46.0       &                   78.0                      &                   74.1                      \\
223         &  --   &   {64.0}    &   --   &   {100}    &  --     &     --         &       --     &   --     &   --  &        {\textbf{82.0}}     &     {\textbf{82.0}} &                      43.8              &                                  \underline{46.6}       &                            45.8             &                   45.8                      \\
255         &   --  &   --    &   --   &   --    &  --     &     {100}         &        --    &     --   &   --  &     {\textbf{100}}        &       {\textbf{100}}  &               \underline{75.0}                   &                                   56.4      &                     \underline{75.0}                 &                      \textbf{100}                \\ 
\midrule mAP &  {100}   &   {76.8}    &    {100}  &   {60.1}    &   {58.0}    &           {75.0}   &      {100}      &    --    &  {80.3}   &      {\textbf{80.8}}       &        {\underline{76.6}} &                 52.4                &                   47.9                      &                    55.1                     &  66.8  \\ \bottomrule
\end{tabular}%
\end{sidewaystable*}

%% file: tables/times.tex
\begin{table}[t!]
\centering
\caption{Processing times of the main stages of the proposed probabilistic framework for instance-aware semantic mapping averaged over the evaluated sequences of the SceneNN~\cite{hua2016_scenenn} dataset. Mask R-CNN is executed on GPU, in parallel to the remaining stages which are assigned to CPU.}
\label{tab:times}
\begin{tabular}{ccc}
\toprule
\textbf{Stage}              & \textbf{Frequency} & \textbf{Time (ms)} \\ \midrule
Mask R-CNN*          &   Per frame  &     69.29   \\
Opinions generation &   Per frame  &     131.28   \\
Data association    &   Per frame  &       9.72    \\
Map integration     &   Per frame  &      19.35     \\
Map refinement      &   Every 30 frames  &   33.19  \\ \midrule
\textbf{Frame-rate} &   &       \textbf{6.24 Hz}     \\ \bottomrule
\end{tabular}%
\end{table}

%% file: tables/Ablation.tex
\begin{table*}[bt]
\centering
\caption{
Ablation studies carried out on the proposed method. All the results correspond to the map obtained directly from Voxeland, without a disambiguation step.
}
\vspace{0.5cm}
\label{tab:ablation}
\begin{tabular}{ccc}
\toprule
\textbf{Ablation}              & \textbf{mAP@0.5} & \textbf{Change from baseline} \\ \midrule
Baseline (w/o disambiguation) &      76.6 & -- \\
\new{w/o Point cloud preprocessing}   &   51.1  &       -33.3\%    \\
\new{w/o Geometric overlap metrics}         &   51.0  &     -33.3\%  \\
\new{w/o Semantic similarity} &   62.0  &    -19.0\% \\
\new{w/o Geometric clustering refinement}   &   69.6  &       -9.1\%    \\
\bottomrule
\end{tabular}%
\end{table*}

%% file: tables/scannet_quantitative.tex
\begin{table}[tb]
\centering
\caption{Quantitative 3D semantic instance-level segmentation performance of Voxeland on the evaluated ScanNet~\cite{dai2017_scannet} sequences. On the left, the Average Precision (AP) is shown for each class (calculated with an Intersection over Union (IoU) threshold $\ge$ 0.5) for Voxeland with disambiguation. The mean Average Precision (mAP) compares our full method with and without disambiguation. A dash indicates the object category is not present in the sequence.}
\label{tab:scannet_eval}
\begin{tabular}{c|cccc||cc}
\toprule
Seq. ID & Sink & Chair & Sofa & Table & \begin{tabular}[c]{@{}c@{}}Voxeland\\(Ours)\end{tabular} & \begin{tabular}[c]{@{}c@{}}Voxeland w/o\\Disambig.\end{tabular} \\ \midrule
scene0027\_00 & -- & 75.0 & 100 & -- &\textbf{87.5} & \textbf{87.5} \\
scene0122\_00 & -- & 100 & -- & 100 & \textbf{100} & 50.0 \\
scene0579\_00 & 100 & -- & -- & -- & \textbf{100} & \textbf{100} \\ \midrule
mAP & 100 & 87.5 & 100 & 100 &\textbf{95.8} & 79.2 \\ \bottomrule
\end{tabular}
\end{table}